\documentclass[journal]{IEEEtran}

\ifCLASSINFOpdf
   \usepackage[pdftex]{graphicx}
   \graphicspath{{../pdf/}{../jpeg/}}
   \DeclareGraphicsExtensions{.pdf,.jpeg,.png}
\else
   \usepackage[dvips]{graphicx}
   \graphicspath{{../eps/}}
   \DeclareGraphicsExtensions{.eps}
\fi
\usepackage{amsmath}
\usepackage{todonotes}
\usepackage{amsfonts}

\usepackage{graphicx} 
\usepackage{epsfig} 

\usepackage{booktabs}
\usepackage{siunitx}
\usepackage{soul,framed}
\usepackage{color}
\usepackage[normalem]{ulem} 

\usepackage{colortbl}
\usepackage{arydshln}
\usepackage[flushleft]{threeparttable}
\usepackage{algorithm,algorithmic}
\usepackage{pifont}
\newcommand{\cmark}{\ding{51}}%
\newcommand{\xmark}{\ding{55}}%

\definecolor{darkgreen}{rgb}{0,0.6,0.2}

\begin{document}
%
\title{SCG-Net: Self-Constructing Graph Neural Networks for Semantic Segmentation}
%
%
%


\author{Qinghui Liu$^{1,2}$,~\IEEEmembership{Student Member,~IEEE,} Michael Kampffmeyer$^{2,1}$,~\IEEEmembership{Member,~IEEE,} Robert Jenssen$^{2,1}$,~\IEEEmembership{Senior Member,~IEEE,} and Arnt-B{\o}rre Salberg$^{1}$,~\IEEEmembership{Member,~IEEE}\\  
\thanks{$^{1}$Norwegian Computing Center, Dept. SAMBA, P.O. Box 114 Blindern, NO-0314 Oslo, Norway}%
\thanks{$^{2}$UiT Machine Learning Group, Department of Physics and Technology, UiT the Arctic University of Norway, Troms{\o}, Norway
        }%
}

\maketitle

\begin{abstract}
Capturing global contextual representations by exploiting long-range pixel-pixel dependencies has shown to improve semantic segmentation performance. However, how to do this efficiently is an open question as current approaches of utilising attention schemes or very deep models to increase the models field of view, result in complex models with large memory consumption. Inspired by recent work on graph neural networks, we propose the Self-Constructing Graph (SCG) module that learns a long-range dependency graph directly from the image and uses it to propagate contextual information efficiently to improve semantic segmentation. The module is optimised via a novel adaptive diagonal enhancement method and a variational lower bound that consists of a customized graph reconstruction term and a Kullback-Leibler divergence regularization term. When incorporated into a neural network (SCG-Net), semantic segmentation is performed in an end-to-end manner and competitive performance (mean F1-scores of $92.0$\% and $89.8\%$ respectively) on the publicly available ISPRS Potsdam and Vaihingen datasets is achieved, with much fewer parameters, and at a lower computational cost compared to related pure convolutional neural network (CNN) based models.
\end{abstract}


%
\IEEEpeerreviewmaketitle

\section{Introduction}
\IEEEPARstart{S}{\lowercase{emantic}} segmentation is one of the fundamental tasks in remote sensing. It refers to classifying each pixel in remote sensing images to a semantic category, e.g., buildings, roads, rivers, etc. Very high resolution (VHR) aerial images often contain diverse objects, highly imbalanced classes, and intricate variations in aspect ratio and color textures (e.g. roads, roofs, shadows of buildings, low plants and branches of trees). This brings complexity and challenges to the semantic segmentation in the field of remote sensing. Modeling the global contextual representations can obtain richer local and non-local information of complex-shaped and context-dependable objects. It has shown to benefit a range of object detection and segmentation tasks~\cite{hu2018relation, zhang2018context}. However, how to efficiently capture long-range dependencies is still an open question for semantic segmentation. In convolutional neural networks (CNNs), the long-range pixel dependencies are mainly modeled by deeply stacked convolutional layers. This results in complex models with a large amount of learnable parameters in particular for the semantic segmentation problem.
The non-local (NL) neural networks~\cite{wang2018non, GCNet} have therefore been proposed to strengthen
the global context modeling ability of CNNs by utilizing a self-attention mechanism~\cite{Attentionisall} within the convolutional structures. The Dual attention network (DANet)~\cite{DANet} further develops two types of NL-style attention modules on top of dilated fully convolutional networks (FCNs)~\cite{long2015fully}, which model the semantic and long-range interdependencies in channel and spatial dimensions respectively. 
However, all of the above attention-based methods have very high memory consumption and can only be embedded into pure CNN-based models. Meanwhile, most existing pure deep CNNs based architectures~\cite{ronneberger2015u, badrinarayanan2017segnet, zhao2016pyramid, wang2017understanding, peng2017large, chen2018deeplab, liu_2019} normally rely on wide multi-scale and multi-streams CNN frameworks to obtain higher performance. This typically requires more trainable parameters and computational resources, often resulting in inefficient and unnecessarily complex models.

Recently, Graph Neural Networks (GNNs) have attracted a lot of attention and have emerged as powerful models to capture global dependencies by leveraging an interaction graph when such a graph is naturally available as a source of information about the problem, such as social networks~\cite{chiang2019cluster, huang2018adaptive}, bio-chemistry~\cite{xu2018powerful, velickovic2019deep}, and so on. Variants of GNNs have also been increasingly explored in various image analysis tasks that include image classification~\cite{knyazev2019image}, few-shot and zero-shot learning \cite{garcia2017few,kampffmeyer2019rethinking, marino2016more}, and have demonstrated very promising performance for various image-level reasoning tasks while significantly reducing the computational cost~\cite{knyazev2019image}. However, GNNs have not yet fully demonstrated their advantages and have been rarely deployed in dense prediction tasks such as semantic segmentation due to lack of prior knowledge graphs.
Previous attempts~\cite{liang2018symbolic, wang2019dynamic, landrieu2018large, qi20173d} either manually produce prior knowledge graphs, or compute other forms of static graph structure based on raw input images, which are neither flexible nor easily generalized to other image datasets. 
Specifically, Wang et al.~\cite{wang2019dynamic} and Landrieu et al.~\cite{landrieu2018large} encoded point clouds into k-nearest neighbor (KNN) graphs or super-point graphs to estimate the global context relations. In Qi et al.~\cite{qi20173d}, a directed graph is constructed based on the 2D position and the depth information of pixels. Pixels close to each other in depth are connected, but those that are close on a 2D grid but distant in depth are not connected with an edge. Note that the prior works mentioned above have not utilized GNNs to infer the final semantic predictions. 
\begin{figure*}[tphb!]
 \centering
  \includegraphics[width=0.90\textwidth]{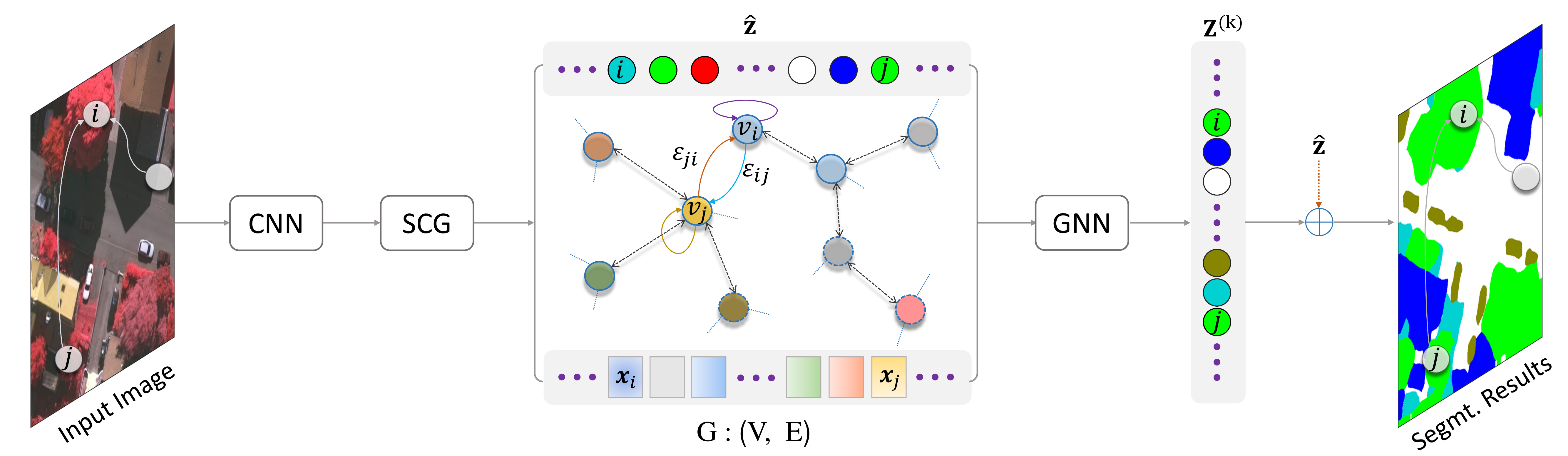} 
      \caption{Our SCG-Net uses a conventional CNN backbone to learn a 2D feature map of an input image. The SCG module then learns to transform the 2D feature map into a latent graph structure $\mathrm{G}:\left(\mathrm{V},\mathrm{E}\right)$, construct the global context relations ($\varepsilon_{ij} \in \mathrm{E}$ ) and assign feature vectors ($\boldsymbol{x}_i$) to the vertices ($v_i \in \mathrm{V}$) of the graph. The $k$-layer GNNs are then exploited to first update the node embedding along the edges of graph with $\left(k-1\right)$ steps and finally predict the node labels, $\mathbf{Z}^{(k)}$, by the $k$-th GNN, the set of node labels are then projected back onto the original 2D plane to output the final segmentation results. }
  \label{fig:overall_scg_net}
\end{figure*}

As a key solution to effectively exploit GNNs to model global representations and long-range context dependencies in remote sensing, we propose a novel self-constructing graph neural network (SCG-Net) model for pixel-level classification tasks as shown in Fig.~\ref{fig:overall_scg_net}. Our SCG-Net model can explicitly employ various kinds of GNNs to not only learn global context representations but also directly output the predictions. More specifically, we introduce a novel Self-Constructing Graph module (SCG), inspired by variational graph auto-encoders (VGAE)~\cite{kipf2016variational} and variational autoencoders (VAE)~\cite{kingma2013auto}, to \emph{learn} how a 2D feature map can be transformed into a latent graph structure and how pixels can be assigned to the vertices of the graph from the available training data. In a nutshell, we model relations between pixels that are spatially similar in the CNN, while in the VAE-based SCG module, we incorporate context information between patches that are similar in feature space, but not necessarily spatially close.

Built upon the proposed SCG module, we further develop the end-to-end SCG-Net model for semantic segmentation tasks. In our SCG-Net framework, a standard CNN (e.g ResNet~\cite{he2016deep}) is utilized to extract high-level feature maps, which are then used to construct the underlying contextual graph using the SCG. A $k$-layer GNN is then exploited to not only learn the latent embedding, but also to infer the final node-wise labels based on the global contextual graph generated by the SCG module. The predicted node labels are finally projected back onto the original 2D plane. Fig.~\ref{fig:overall_scg_net} presents an overview of our method. 

Compared to most previous work on graph reasoning for scene recognition tasks, our SCG-Net effectively streamlines the semantic segmentation pipeline by transforming the pixel-level classification problem in an Euclidean domain into a node-wise classification task in a structural domain, without relying on deep and multi-scale feature fusion architectures. The proposed SCG framework provides flexibility as it seamlessly combines CNNs with variants of GNNs (e.g.~ \cite{gori2005new, hammond2011wavelets, bronstein2017geometric, niepert2016learning, defferrard2016convolutional, kipf2016semi, hamilton2017inductive, xu2018powerful}) together to solve semantic segmentation problems.
Our experiments demonstrate that SCG-Net achieves very competitive accuracy with real-time performance on the representative ISPRS 2D semantic labeling datasets~\cite{ISPRS2018}. In summary, our contributions are: 
\begin{enumerate}
  \item We propose a new end-to-end architecture for semantic segmentation tasks, namely the  self-constructing graph neural network (SCG-Net), that successfully simplifies the image segmentation pipeline.
  \item The proposed SCG module can be easily integrated to existing CNNs and GNNs architectures 
  to leverage the benefits of both GNN and CNN. 
  \item Our proposed end-to-end model achieves competitive performance on different representative remote sensing datasets with much fewer parameters, faster training and lower computational cost.
\end{enumerate}

A preliminary version of this paper appeared in \cite{liu2020scg, Liu_2020_CVPR_Workshops}. Here, we extend our work by (i) extending our method with several new variants such as Autoencoder (AE) based SCG, Directed SCG, and various combinations with different GNNs, to further boost the model's flexibility;
(ii) expanding the experiment section by including more datasets with more variants of models, providing more  training details and presenting additional result comparisons and analysis;
(iii) providing sound ablation studies, qualitative analysis 
and in-depth discussions in terms of model's effectiveness, limitations and challenges for future work.

The paper is structured as follows. Section~\ref{prel} introduces some preliminaries and background.  In Section~\ref{method}, we present the methodology in details. Section~\ref{data} introduces the datasets used in our work. Experimental procedure and evaluation of the proposed method is performed in Section~\ref{exp}. And, finally in Section~\ref{concl}, we draw conclusions.

\section{Preliminaries}
\label{prel}
\begin{table}[htbp]
  \centering
  \caption{Commonly used notations in this paper.}
  \resizebox{\columnwidth}{!}{
    \begin{tabular}{c|c}
    \textbf{Notations} & \textbf{Descriptions} \\
    \midrule
    $\mathrm{G}$ & A graph \\ \hline
    $\mathrm{V}$ & The set of nodes (vertices) in a graph  \\ \hline
    $\mathrm{E}$ & The set of edges (pairs/links of nodes) in a graph \\ \hline
    $\mathrm{C}$ & The label set $\{1,2,\dotsc, c\}$ \\ \hline
    $\mathrm{Y}$ & The set of ground truth for all labeled nodes, composed by $\{y_i\}$  \\ \hline
    $v_i$ & The $i$-th node  $\in \mathrm{V}$  \\ \hline
    $\mathrm{N}_{v_i}$ & The set of nodes adjacent to $v_i$  \\ \hline
    $\varepsilon_{ij}$ & The link $\in \mathrm{E}$ of the node pair $(v_i,v_j)$ directed from $v_i$ to $v_j$  \\ \hline
    $\boldsymbol{x}_i \in \mathbb{R}^{d}$ & The $d$-dimensional feature vector associated to $v_i$\\ \hline
    $\mathbf{A}\in \mathbb{R}^{n \times n}$ & The adjacency matrix of a graph \\ \hline
    $\mathbf{D}\in \mathbb{R}^{n \times n}$ & The degree matrix of $\mathbf{A}$ with self-loop \\ \hline
    $\mathbf{\hat{A}}\in \mathbb{R}^{n \times n}$ & The normalized graph adjacency matrix \\ \hline
    $\mathbf{X}\in \mathbb{R}^{n \times d}$ & The feature matrix of a graph, composed by $[\boldsymbol{x}_i]$ \\ \hline
    $\mathbf{\hat{Z}} \in \mathbb{R}^{n \times c}$ & The residual predictions produced by SCG module \\ \hline
    $\mathbf{Z}^{(k)} \in \mathbb{R}^{n \times c}$ & The predictions produced by GNN module \\ \hline
    $\mathbf{I}$ & The identity matrix \\ \hline
    $\boldsymbol{\theta}^{(l)}$ & The learnable parameters of the $l\text{-th}$ layer \\ \hline
    $\mathbf{Z}^{(l)}$ & The hidden feature of the $(l)\text{-th}$ layer, composed by $[\boldsymbol{z}^{(l)}_i]$ \\ \hline
    $\omega^{(l)}$ & A learnable parameter or a fixed scalar at layer $l$ \\ \hline
    $\boldsymbol{z}^{(l)}_i$ &  The hidden feature vector of node $v_i$ at the $l$-th layer \\ \hline
    $n$ & The number of nodes \\ \hline
    $d$ & The dimension of a node feature vector \\ \hline
    $c$ & The number of classes \\ \hline
    $k$ & The last layer index  \\ \hline
    $l$ & The hidden layer index  \\ \hline
    $y_i$ & The true label of node $v_i$  \\ \hline
    $\delta\left(\cdot\right)$ & The activation function, such as $\operatorname{ReLU}$  \\ \hline
    $\left|{\cdot}\right|$ & The length of a set  \\ 
    \bottomrule
    \end{tabular}%
  \label{tab:notations}%
  }
\end{table}%
Here we begin by presenting some relevant background of the most common GNN models. The notations used in this paper are shown in Table~\ref{tab:notations}. Unless particularly specified, we use upright letters to denote sets and subsets, bold capital characters for matrices, lowercase in italics for scalars and bold italics for vectors.
\subsection{Graph neural networks for node classification}
Consider a graph $\mathrm{G}=\left(\mathrm{V},\mathrm{E}\right)$ that consists of a set $\mathrm{V}=\{v_i=(i, \boldsymbol{x}_i): i=1,2,\dotsc, n\}$ of $n$ vertices or nodes,  where $\boldsymbol{x}_i$ denotes the $d$-dimensional feature vectors for node $v_i$, and an associated set of edges $\mathrm{E} = \{\varepsilon_{ij}=(i, j, A_{ij}): i=1,2,\dotsc, n,\, j=1,2,\dotsc, n\}$, where $A_{ij}$ represents the weight associated to the node pair $(v_i,v_j)$ directed from $v_i$ to $v_j$. Here, we assume a set of labeled nodes $\{(v_i, y_i \in \mathrm{Y}): i= 1,2,\dotsc ,m\}$, where $\mathrm{Y}$ contains the ground truth for all the labeled nodes from a label set $\mathrm{C} = \{1,2,\dotsc, c\}$, and $m = \left|{\mathrm{Y}}\right| \leq n$. The goal of node classification problem is to learn a mapping $f: \mathrm{V} \rightarrow \mathrm{C}$ such that the labels of unlabeled nodes can be predicted.


For simplicity, $\mathrm{G}$ can be also represented by ($\textbf{A}, \textbf{X}$), where\footnote{Note that we assume $\mathrm{G}$ is a weighted graph instead of binary one in this paper.}, the adjacency matrix $\textbf{A} \in \mathbb{R}^{n \times n}$ is composed of each link weight $A_{ij} \geq 0 \in \mathbb{R}$, and the feature matrix $\textbf{X} \in \mathbb{R}^{n \times d}$ contains each node embedding $\boldsymbol{x}_i \in \mathbb{R}^{d}$. Essentially, a GNN can be viewed as a "Message-Propagating" (MP) operation in graph structures, since it learns latent features, $\mathbf{Z}^{(l)}$, by recursively aggregating the information from neighbouring nodes in the graph. The generalized message-propagation architecture can be defined as
\begin{equation} \label{eq:message}
\mathbf{Z}^{(l)} = \mathrm{MP}(\mathbf{\hat{A}}, \mathbf{Z}^{(l-1)}; \boldsymbol{\theta}^{(l)}), l=1,2,\dotsc, k\;, 
\end{equation}
where $\mathbf{Z}^{(l-1)}$ denotes the node features at the $(l-1)\text{-th}$ layer and $\mathbf{Z}^{(0)} = \mathbf{X}$, $\boldsymbol{\theta}^{(l)}$ are the trainable parameters of the $l\text{-th}$ layer, $\mathbf{Z}^{(l)}$ is the latent embedding space computed after $(l)$ steps and $\mathrm{MP}(\cdot )$ denotes the message-propagation function. Note that $\mathbf{A}$ is often re-normalized in a particular way to a normalized matrix $\mathbf{\hat{A}}$ based on the specific GNN variant~\cite{kipf2016semi, xu2018powerful}. 

The most common GNNs follow the message-propagation strategy that can be generalized as Eq.~\ref{eq:message}. There are many kinds of implementations of the propagation function. In this work, we mainly exploit two representative GNN variants, namely the spectral-based method - Graph Convolutional Network (GCN)~\cite{kipf2016semi}, and the spatial-based method - Graph Isomorphism Network (GIN)~\cite{xu2018powerful}.

\subsection{Graph Convolutional Network}
The GCN~\cite{kipf2016semi} was presented as the first-order approximation of the spectral GNN~\cite{hammond2011wavelets}, that implements a message-propagation function by a combination of linear transformations over one-hop neighbourhoods and non-linearities. It is defined as
\begin{equation} \label{eq:gcn2}
\mathbf{Z}^{(l)}=\delta \left(\mathbf{\hat{A}} \mathbf{Z}^{(l-1)}, \boldsymbol{\theta}^{(l)}\right)\;,
\end{equation}
where $\delta$ denotes the non-linearity function (e.g. $\operatorname{ReLU}$), $\mathbf{\hat{A}}$ is the normalized version of $\mathbf{A}$ with self-loops given as

\begin{equation} \label{eq:norm}
\mathbf{\hat{A}}=\mathbf{D}^{-\frac{1}{2}}(\mathbf{A}+\mathbf{I})\mathbf{D}^{\frac{1}{2}}\;,
\end{equation}
where $D_{ii} = \sum_{j}(\mathbf{A}+\mathbf{I})_{ij}$ is a diagonal matrix with node degrees, and $\mathbf{I}$ is the identity matrix. A complete description of the GCN can be found in~\cite{kipf2016semi}.

\subsection{Graph Isomorphism Network}
Unlike the spectral-based GNN, GIN~\cite{xu2018powerful} was proposed as a spatial-based method that updates the node embedding based on the spatial relations of vertices. More specifically, the GIN's message-propagating function can be defined as

\begin{equation} \label{eq:gin1}
\boldsymbol{z}_{i}^{(l)}=\text{MLP}^{(l)}\left(\left(1+\omega^{(l)}\right) \cdot \boldsymbol{z}_{i}^{(l-1)}+\sum_{j\in \mathrm{N}_{v_i}} \boldsymbol{z}_{j}^{(l-1)}\right)\;,
\end{equation}
where $\boldsymbol{z}_{i}^{(l)} \in \mathbf{Z}^{(l)}$ is the feature vector of node $v_i$ at the $l$-th iteration, $\mathrm{N}_{v_i}$ represents an set of nodes adjacent to $v_i$, $\text{MLP}^{(l)}$ denotes the multi-layer-perception (MLP) at layer $l$, and $\omega^{(l)}$ is a learnable parameter or a fixed scalar at layer $l$. 
Eq.~\ref{eq:gin1} can be converted to a dense/matrix representation as 

\begin{equation} \label{eq:gin2}
\mathbf{Z}^{(l)}=\delta\left(\left(\left(1 +\omega^{(l)}\right)\mathbf{I} + \mathbf{A}\right)\mathbf{Z}^{(l-1)} \boldsymbol{\theta}^{(l)}\right) \;.
\end{equation}
Comparing Eq.~\ref{eq:gin2} to Eq.~\ref{eq:gcn2}, the major difference between the GIN and the GCN is that the normalized adjacency matrix $\mathbf{\hat{A}}$ is replaced by $\left(\omega^{(l)}\mathbf{I} + (\mathbf{I} + \mathbf{A})\right)$. We can therefore consider the GIN as a special version of the GCN that takes the raw adjacency matrix with a learnable or fixed-scaled diagonal matrix rather than using a Laplacian normalized one for message-propagation. For more details about the GIN, please refer to the work~\cite{xu2018powerful}.

In the following sections, we will use $\mathbf{Z}^{(k)} = \operatorname{GNN}(\mathbf{A}, \mathbf{X})$ to denote an arbitrary $\operatorname{GNN}$ module (e.g., GCN or GIN in this paper) implementing $k$ steps of message passing based on some adjacency matrix $A$ and input node features $X$, where $k$ is commonly in the range $2$ to $6$.

\section{The SCG-Net}
\label{method}
We first describe the proposed Self-Constructing Graph (SCG) framework and its variants. We then present our design choices and provide detailed information about the end-to-end model SCG-Net for semantic segmentation tasks.

\begin{figure*}[htbp]
 \centering
  \includegraphics[width=0.70\textwidth]{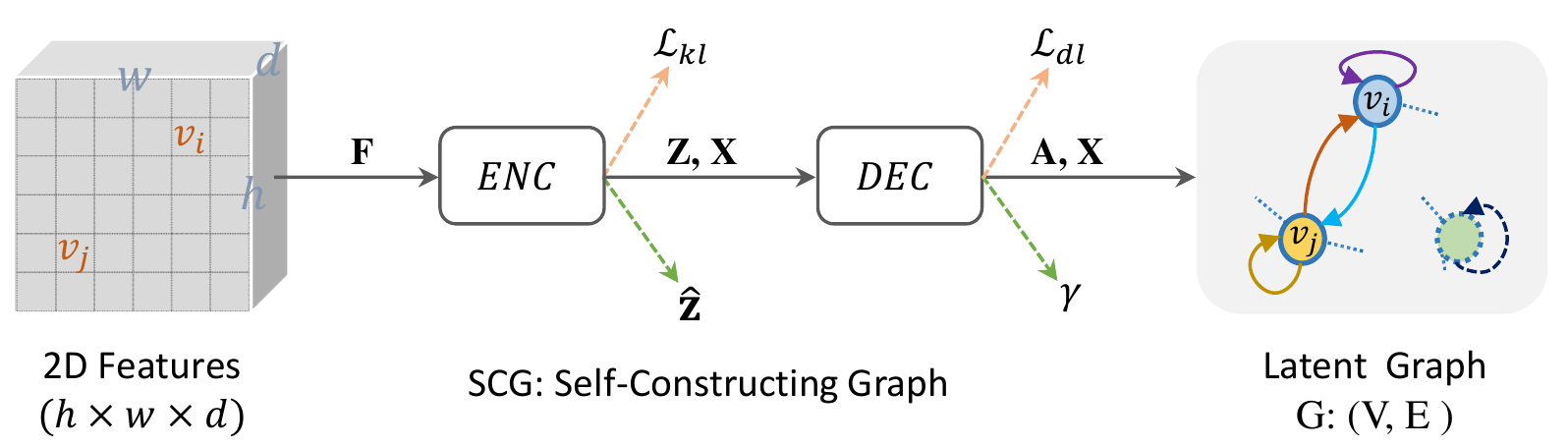} 
  \caption{The illustration diagram of the SCG model. Overall, it is composed of 2 key modules, i.e., the \textit{ENC} module that transforms the 2D features, $\textbf{F} \in \mathbb{R}^{h \times w \times d}$, to a latent embedding space, $\mathbf{Z} \in \mathbb{R}^{n \times c}$, and, the \textit{DEC} module that generate the graph representations $\mathrm{G}:(\mathbf{A}, \mathbf{X})\Leftrightarrow(\mathrm{V},\mathrm{E}) $ by decoding the learned latent embeddings from \textit{ENC}. Note that SCG model also introduce two regularization functions, i.e., $\mathcal{L}_{kl}$ is the Kullback–Leibler divergence loss and $\mathcal{L}_{dl}$ is the diagonal log loss, and two auxiliary terms, i.e.,  $\gamma$ is the adaptive factor, $\mathbf{\hat{Z}}$ is the residual embeddings/predictions.}
  \label{fig:scg_module}
\end{figure*}

\subsection{The Framework of SCG}
We propose the SCG model that aims to learn the latent graph representation for capturing the global context information across the scene image directly from 2D feature maps without relying on prior knowledge. Overall, the framework of SCG contains 2 key modules, i.e the \textit{ENC} module that transforms the input 2D features, $\textbf{F} \in \mathbb{R}^{h \times w \times d}$, 
to a latent embedding space, $\mathbf{Z} \in \mathbb{R}^{n \times c}$, and, the \textit{DEC} module that generates the graph representations $\mathrm{G}:(\mathrm{V},\mathrm{E}) \Leftrightarrow (\mathbf{A}, \mathbf{X})$ by decoding the learned latent embeddings from \textit{ENC} (see Figure~\ref{fig:scg_module}). 
Additionally, to further improve the learning process, the SCG module introduces two regularization terms, i.e the Kullback-Leibler divergence term ($\mathcal{L}_{kl}$) and the diagonal log penalty ($\mathcal{L}_{dl}$), and two adaptive enhancement methods, i.e the adaptive enhancement factor ($\gamma$) with the residual embeddings ($\mathbf{\hat{Z}}$) to refine the final predictions. Hence, the generalized SCG function can be defined as 
\begin{equation} \label{eq:scg}
(\text{G}, \mathcal{L}_{kl}, \mathcal{L}_{dl}, \mathbf{\hat{Z}}, \gamma) = \operatorname{SCG}(\textbf{F})\;.
\end{equation}
Concretely, we describe each component of the SCG in details as follows.

\subsubsection{\textbf{ENC}}
In the ENCoding (ENC) module, we first take a parameter-free pooling operation (e.g. adaptive average pooling in our case) to reduce the spatial dimensions of $\textbf{F}$ 
from $(h \times w)$ to  $(\overline{h} \times \overline{w})$ of $\overline{\textbf{F}} \in \mathbb{R}^{\overline{h} \times \overline{w} \times d}$, and then flatten $\overline{\textbf{F}}$ to obtain $X$ as the 1-D feature matrix containing $n$ ($n = \overline{h} \times \overline{w}$) nodes.  
Meanwhile, the \textit{ENC} module, following convention with variational autoencoders (VAE)~\cite{kingma2013auto}, learns a mean matrix $\textbf{M} \in \mathbb{R}^{n \times c}$
and a standard deviation matrix $\boldsymbol{\Sigma} \in \mathbb{R}^{n \times c}$ of a Gaussian using two single-layer convolutional networks 
\begin{equation}\label{eq:u}
    \textbf{M}  = \operatorname{Flatten}(\operatorname{Conv}_{3 \times 3}(\overline{\textbf{F}}))
\end{equation}
\begin{equation}\label{eq:var}
    \log(\boldsymbol{\Sigma}) = \operatorname{Flatten}\left(\operatorname{Conv}_{1 \times 1}(\overline{\textbf{F}})\right),
\end{equation}
where $c$ denotes the number of classes and the index (i.e., $_{3\times3}$ and $_{1\times1}$) specifies the kernel size of the convolutional layer.
Note that the output of the model for the standard deviation is $\log(\boldsymbol{\Sigma})$ to ensure stable training and positive values for $\boldsymbol{\Sigma}$. 

With help of reparameterization~\cite{kingma2013auto}, the latent embedding $\mathbf{Z}$ is obtained as $\mathbf{Z} = \textbf{M} + \boldsymbol{\Sigma} \cdot  \boldsymbol{\Upsilon}$ \;
where $\boldsymbol{\Upsilon} \in \mathbb{R}^{n \times c}$ is an auxiliary noise variable and initialized from a standard normal distribution ($\boldsymbol{\Upsilon} \sim \boldsymbol{\mathcal{N}}_{n\times c}\left(\textbf{0},\textbf{I}\right)$). A centered isotropic multivariate Gaussian prior distribution is used to regularize the latent variables during training, by minimizing a Kullback-Leibler divergence loss~\cite{kingma2013auto} that is given as

\begin{equation}\label{eq:klloss}
\mathcal{L}_{kl} = -\frac{1}{2nc} \sum_{i=1}^{n}\sum_{j=1}^{c}\left(1+\log \left({\it \Sigma}_{ij} \right)^{2}-{M}_{ij}^{2}-\left({\it \Sigma}_{ij}\right)^{2}\right) \;.
\end{equation}

Here, we also introduce a residual embedding term $\mathbf{\hat{Z}}$ that is defined as: $\mathbf{\hat{Z}} = \textbf{M} \cdot (1 - {\log \boldsymbol{\Sigma}})$, which can be used to refine the final prediction of the network after information has been propagated along the learned graph. Intuitively, the residual term $\mathbf{\hat{Z}}$ can be viewed as a standard normal distribution space transformed from the latent space by $\textbf{M} \cdot \boldsymbol{\Sigma}^{-1}$. 
For computational simplicity and stabilization, we replace $\boldsymbol{\Sigma}^{-1}$ with $(1 - {\log \boldsymbol{\Sigma}})$, and $\log \boldsymbol{\Sigma}$ is constrained to be smaller than 1 during the training. 
\subsubsection{\textbf{DEC}}
In the DEC-block, the graph adjacency matrix $\mathbf{A}$ is generated by an inner product between latent embeddings as $\mathbf{A} = \operatorname{ReLU}(\mathbf{Z} \mathbf{Z}^{T})$. Note, $\mathbf{A}$ is not binary in our case but weighted and undirected. Thus, $A_{ij} = A_{ji} > 0 $ indicates the presence of an edge associated with the learned weight between node $i$ and $j$. Essentially, the DEC-block measures the similarity between patches (feature vectors of nodes) to build the graph that connects similar node representations together, such that the similar scene regions exchange information. And intuitively, we consider $A_{ii}$ shall be $>0$ and close to $1$. We therefore introduce a diagonal log regularization term defined as
\begin{equation}\label{eq:dlloss}
  \mathcal{L}_{dl} = - \frac{\gamma}{n^2} \sum_{i=1}^{n} \operatorname{log} (\left|A_{ii}\right|_{[0,1]} + \epsilon)  \;,
\end{equation}
where the index $_{[0,1]}$ denotes that $A_{ii}$ is clamped to $[0, 1]$, $\epsilon$ is a small positive infinitesimal scalar (e.g, $\epsilon = 10^{-7}$) and $\gamma$ is the adaptive factor computed as $$\gamma = \sqrt{1 + \frac{n}{ \sum_{i=1}^{n} A_{ii} + \epsilon}}\;.$$

In order to preserve local information and stabilize training, we introduce an adaptive enhancement approach applied to both the adjacency matrix and the residual embeddings. The enhanced $\mathbf{A}$ is designed as $\mathbf{A} = \left(\mathbf{A} + \gamma \cdot \operatorname{diag}(\mathbf{A})\right)$, and the enhanced residual term $\mathbf{\hat{Z}}$ is defined as $\mathbf{\hat{Z}} = \gamma \cdot \mathbf{\hat{Z}}$. 




\subsection{Variants of SCG}
\subsubsection{Auto-encoder based SCG}
We introduce an auto-encoder based SCG module ($\operatorname{SCG}_{ae}$) instead of using a variational auto-encoder based structure to learn the latent embedding $Z$. The ENC module can thus learn $Z$ as follows
\begin{equation}\label{eq:ae-z}
    \mathbf{Z}  = \operatorname{Flatten}(\operatorname{Conv}_{3 \times 3}(\overline{\textbf{F}}))\;.
\end{equation}
Let the residual term be $\mathbf{\hat{Z}} = \mathbf{Z}$. Then the overall formula of $\operatorname{SCG}_{ae}$ simplifies to 
\begin{equation} \label{eq:scg-ae}
(\text{G}, \mathcal{L}_{kl}, \mathbf{\hat{Z}}, \gamma) = \operatorname{SCG}_{ae}(\textbf{F})\;.
\end{equation}
Note that there is no need to compute the mean matrix ($\textbf{M}$), the deviation matrix ($\boldsymbol{\Sigma}$), and the Kullback-Leibler divergence loss ($\mathcal{L}_{kl}$) any more. However, the DEC part of $\operatorname{SCG}_{ae}$ is kept unchanged.

\subsubsection{Directed graph-based SCG}
The standard $\operatorname{SCG}$ framework can only generate un-directed graphs (symmetric adjacency matrix) by the inner product operation. Here we propose a SCG variant $\text{SCG}_{dir}$ that is able to produce directed graph structures. To do so, we first normalize the latent embeddings by the standard softmax operation as 
\begin{equation}\label{eq:softamx}
    \Bar{Z}_{ij}=\frac{e^{Z_{ij}}}{\sum_{k=1}^{c} e^{Z_{ik}}} \quad i= 1, 2, \dotsc, n \quad  j=1,2,\dotsc, c \;.
\end{equation}
Then the directed adjacency matrix can be defined as $\mathbf{A} = \operatorname{ReLU}(\mathbf{\Bar{Z}} \mathbf{Z}^{T})$. Note that $A_{ij} \neq A_{ji} > 0$ indicates the presence of directed edges associated with two different weights between node $i$ and $j$. Intuitively, we assume that the directed graph can represent richer interactions between objects. 


\subsection{SCG-Net Architecture} \label{archit}
The SCG module can be easily incorporated into existing CNN and GNN architectures from which a more complex network can be built. Fig.~\ref{fig:overall_scg_net} shows the illustration of the proposed SCG-Net model for semantic segmentation tasks. The model architecture details are shown in Table~\ref{tab:SCG-Net}.

\begin{table}[htbp]
  \centering
  \caption{The general end-to-end SCG-Net Model Details with one sample of input image size of $h_0\times w_0 \times 3$.}
  \resizebox{\columnwidth}{!}{
    \begin{tabular}{c|c|c}
    \textbf{Layers} & \textbf{Outputs} & \textbf{Sizes} \\
    \midrule
    Backbone-CNN & $\textbf{F}$ & $h\times w\times 1024$  \\ \hdashline
    SCG & $(\mathbf{A}, \mathbf{X}, \mathbf{\hat{Z}}, \gamma)$ &  $(n\times n), (n\times 1024), (n\times c), \mathbb{R}_{\ge 1}$  \\ 
    $\operatorname{GNN}^1$ & ($\mathbf{\hat{A}}$, $\mathbf{Z}^{(1)}$) &   $n\times d$      \\ \hdashline
    $\operatorname{GNN}^2$ & ($\mathbf{\hat{A}}$, $\mathbf{Z}^{(2)}$) &   $n\times c$      \\ 
    Sum(opt) & $(\gamma \cdot \mathbf{\hat{Z}} + \mathbf{Z}^{(2)})$ &   $n \times c$  \\ \hdashline
    Projection & $\mathbf{\Tilde{Y}}$ &   $h_0\times w_0\times c$  \\
    \bottomrule
    \end{tabular}%
  \label{tab:SCG-Net}%
  }
\end{table}%

\subsubsection{CNN backbone}
A conventional CNN backbone, e.g, the pretrained ResNet50~\cite{he2016deep} in this work, is employed to extract the high-level representations. Following our previous work~\cite{liu_2019}, we utilize the first three bottleneck layers of ResNet50. We assume that the size of the input image is $h_0\times w_0\times 3$ (with $3$ color channels), the output size of our CNN backbone is thus $h\times w\times 1024$ ($h, w = \frac{h_0}{16}, \frac{w_0}{16}$).
\subsubsection{SCG and GNN decoder}
We combine our proposed SCG module (e.g, the standard $\operatorname{SCG}$, or other variants including $\operatorname{SCG}_{ae}$ and $\operatorname{SCG}_{dir}$) with a 1-layer GNN (either GCN or GIN in this work) as the decoder. We utilize ReLU activation and batch normalization for the GNN layer. Guided by our empirical observation, we set $n= h \times w$ and $d=128$ in this work, and $c=6$ is equal to the number of classes in the datasets.

\subsubsection{Prediction and projection}
The final prediction ($\mathbf{Z}^{(2)}$) is produced by the second layer GNN (either GCN or GIN) without activation and normalization. 
There is also an optional element-wise sum operation for the residual term ($\gamma \cdot \mathbf{\hat{Z}}$) to refine the final results. To project the representations back to the 2D space , we first reshape the predictions ($n \times c \longrightarrow h \times w \times c$), and then conduct up-sampling with bilinear interpolation to obtain the final segmentation maps with original spatial resolution size ($h_0 \times w_0 \times c$).

\begin{table}[htbp]
  \centering
  \caption{Variants of SCG-Net Models with different settings for evaluations in this paper.}
  \resizebox{\columnwidth}{!}{
    \begin{tabular}{c|c|c|c|c}
    SCG-Net Variants & SCG Layer & $\operatorname{GNN}^1$ & $\operatorname{GNN}^2$  & Sum(opt)\\
    \midrule
    $\operatorname{SCG-GCN}$  & $\operatorname{SCG}$ & $\operatorname{GCN}$ & $\operatorname{GCN}$ & \cmark \\ 
     $\operatorname{SCG-GIN}$   & $\operatorname{SCG}$ & $\operatorname{GIN}$ & $\operatorname{GIN}$ &\cmark \\ 
     $\operatorname{SCG-GCN-GIN}$   & $\operatorname{SCG}$ & $\operatorname{GCN}$ & $\operatorname{GIN}$ &\cmark \\ 
    $\operatorname{SCG_{ae}-GCN}$ & $\operatorname{SCG}_{ae}$ & $\operatorname{GCN}$ & $\operatorname{GCN}$  & \cmark\\ 
    $\operatorname{SCG_{dir}-GCN}$   & $\operatorname{SCG}_{dir}$ & $\operatorname{GCN}$  & $\operatorname{GCN}$ & \cmark\\ 
    $\operatorname{SCG_{dir}-GCN-GIN}$   & $\operatorname{SCG}_{dir}$ & $\operatorname{GCN}$  & $\operatorname{GIN}$ & \cmark\\ 
    $\operatorname{SCG-GCN}_{\_ns}$  & $\operatorname{SCG}$ & $\operatorname{GCN}$ & $\operatorname{GCN}$ & \xmark \\ 
    \bottomrule
    \end{tabular}%
  \label{tab:models}%
  }
\end{table}%
\subsubsection{Variants of SCG-Net models}
Table~\ref{tab:models} shows some variants of the SCG-Net model that were investigated in this work. Note that we utilize the same backbone CNN (customized ResNet50) for all these models.
Our proposed SCG-Net framework can be easily extended and tailored to various deep CNNs and GNN-like networks with flexible configurations w.r.t the depth (e.g. the number of layers), width (e.g. the size of inputs) and density (e.g. the number of nodes and hidden features) for different problems. 


\section{Benchmark Datasets}
\label{data}
We evaluate our proposed methods on two public benchmark datasets, namely the ISPRS 2D semantic labeling contest datasets~\cite{ISPRS2018}.
The ISPRS datasets are comprised of aerial images over two cities in Germany: Potsdam\footnote{http://www2.isprs.org/commissions/comm3/wg4/2d-sem-label-potsdam.html} and Vaihingen\footnote{http://www2.isprs.org/commissions/comm3/wg4/2d-sem-label-vaihingen.html}. They have been labelled with six common land cover classes: impervious surfaces, buildings, low vegetation, trees, cars and clutter. 

\subsubsection{Potsdam}
The Potsdam dataset consists of 38 tiles of size $6000 \times 6000$ pixels with a ground resolution of 5cm. 14 of these are used as hold-out test images. Tiles consist of Red-Green-Blue-Infrared (RGB-IR) four-channel images. While both the digital surface model (DSM) and normalized DSM (nDSM) data are also included in the dataset, we only use RGB images in this paper in order to fairly compare with other work. 

\subsubsection{Vaihingen}
The Vaihingen dataset contains 33 tiles of varying size (on average approximately $2100 \times 2100$ pixels) with a ground resolution of 9cm, of which 17 are used as hold-out test images. Tiles are composed of Infrared-Red-Green (IRRG) 3-channel images. Though DSMs and nDSMs data are also available for all images in the dataset, we only focus on the 3-channel IRRG data in this paper to fairly compare with other work. 

\section{Experiments and Results}
\label{exp}
In this section, we investigate our proposed methods and networks on the Potsdam and Vaihingen (Section~\ref{exp-pots}) datasets. We first present the training and evaluation details, and then report both qualitative and quantitative results for the multi-class semantic labeling task.

\subsection{Training details}
Following our previous work~\cite{liuqinghui2018}, we train the models using Adam \cite{KingmaB14adam} with AMSGrad~\cite{amsgrad2018} as the optimizer with weight decay $2 \times 10^{-5}$ applied to all learnable parameters except biases and batch-norm parameters, and polynomial learning rate (LR) decay $(1 - \frac{cur\_iter}{max \_iter})^{0.9}$ with the maximum iterations of $10^8$. The learning rate of the bias parameters is $2 \times \text{LR}$. We use initial LRs of $\frac{8.5 \times 10^{-5}}{\sqrt{2}}$ and utilized a step-wise LR schedule method that reduces the LR by a factor of 0.85 every 15 epochs. Base on our training observations to achieve fast and stable convergence, we apply a dice loss function~\cite{milletari2016v} $\mathcal{L}_{dice}$ defined as

\begin{equation}\label{eq:dice_loss}
\mathcal{L}_{dice} = 1 - \frac{1}{\left|{\mathrm{Y}}\right|}\sum_{i \in \mathrm{Y}} \frac{2 \sum_{j \in \mathrm{C}} y_{ij} \Tilde{y}_{ij}}{\sum_{j \in \mathrm{C}} y_{ij} + \sum_{j \in \mathrm{C}} \Tilde{y}_{ij}}\;,
\end{equation}
where $\mathrm{Y}$ contains all the labeled nodes, $\mathrm{C}$ denotes the label set, and $\Tilde{y}_{ij}$ is the $i.j$-th node with the ground-truth label to be $y_{ij}$.

Together with the two regularization terms $\mathcal{L}_{kl}$ and $\mathcal{L}_{dl}$ as defined in Eq.~\ref{eq:klloss} and Eq.~\ref{eq:dlloss}, respectively, the final cost function of our model is defined as
\begin{equation}\label{eq:cost_function}
    \mathcal{L} = \mathcal{L}_{dice} + \mathcal{L}_{kl} + \mathcal{L}_{dl}\;.
\end{equation}

We train and validate the networks for both the Potsdam and Vaihingen datasets with 4000 randomly sampled patches of size $448\times 448$ as input and using a batch size of $4$. The training data is sampled uniformly and randomly shuffled for each epoch. We conduct all experiments in this paper using PyTorch \cite{paszke2017automatic} on a single computer with one NVIDIA 1080Ti GPU. 

\subsection{Augmentation and evaluation methods}
We randomly flip or mirror images for data augmentation (with probability $0.5$). The albumentations library~\cite{Albumentations2018} for data augmentation is utilized in this work. Please note that all training images are normalized to [0.0, 1.0] after data augmentation.

Following previous work~\cite{liu_2019}, we apply test time augmentation (TTA) via flipping and mirroring. We also use sliding windows (with $448 \times 448$ size at a $100$-pixel stride) on a test image and stitch the results together by averaging the predictions of the over-lapping TTA regions to form the output. The performance is measured by both the F1-score, and the mean Intersection over Union (IoU).

\subsection{Test results} \label{exp-pots}
For evaluation, the labeled part of the Potsdam dataset is split into a training set (19 images), a validation set (2 images of 4\_10 and 7\_10), and a local test set (3 images of areas 5\_11, 6\_9 and 7\_11). The Vaihingen dataset is similarly divided into training (10 images), validation (2 images of areas 7 and 9) and local test set (4 images of areas 5, 15, 21 and 30). While the hold-out test sets contain 14 images (areas: 2\_13, 2\_14, 3\_13, 3\_14, 4\_13, 4\_14, 4\_15, 5\_13, 5\_14, 5\_15, 6\_13, 6\_14, 6\_15 and 7\_13) and 17 images (areas: 2, 4, 6, 8, 10, 12, 14, 16, 20, 22, 24, 27, 29, 31, 33, 35 and 38) for the Potsdam and Vaihingen datasets, respectively.
\begin{table}[hptb!]
\centering 
  \caption{Comparisons between our method with other published methods on the hold-out RGB test images of Potsdam dataset.}
\resizebox{\columnwidth}{!}{
\begin{threeparttable}
\begin{tabular}{c|p{9mm}|p{9mm}p{8mm}p{11mm}p{9mm}p{9mm}|p{8mm}} \hline
    \textbf{Models} & $\textbf{OA}$ & \textbf{Surface} &\textbf{Building} & \textbf{Low-veg} & \textbf{Tree} &  \textbf{Car} & \textbf{mF1} \\  \hline  \hline
    HED+SEG.H-Sc1~\cite{MarmanisSWGDS16} & 0.851  & 0.850  & 0.967  & 0.842  & 0.686  & 0.858  & 0.846 \\  
    RGB+I-ensemble~\cite{michael2018}    & 0.900  & 0.870  & 0.936  & 0.822  & 0.845  & 0.892 & 0.873 \\
    Hallucination~\cite{michael2018}     & 0.901  & 0.873  & 0.938  & 0.821   & 0.848  & 0.882  & 0.872 \\
    DNN\_HCRF~\cite{liu2019semantic}     & 0.884  & 0.912  & 0.946  & 0.851 & 0.851 & 0.928  & 0.898\\ 
    SegNet RGB~\cite{audebert2017joint}  & 0.897  & 0.930  & 0.929  & 0.850  & 0.851  & 0.951  & 0.902\\ 
    DST\_2~\cite{Sherrah16}              & 0.903  & 0.925  & 0.964  & 0.867 & 0.880  & 0.947 & 0.917 \\
    FuseNet+OSM~\cite{audebert2017joint} &  \textbf{0.923}  & \textbf{0.953}  & 0.959  & 0.863 & 0.851  & \textbf{0.968}  & 0.918\\ 
    DDCM-R50~\cite{liu_2019} & 0.908  & 0.929  & \textbf{0.969}   & \textbf{0.877}   &\textbf{0.894}  & 0.949  & \textbf{0.923}\\ \hdashline
     $\textbf{SCG-GCN}$ & 0.903  & 0.924  & 0.952  & 0.873 & 0.893  & \textbf{0.960}  & 0.920 \\  \hline
\end{tabular}
  \end{threeparttable}
} 
\label{tab:potsdam_scores}%

\end{table}
\begin{table}[hptb!]
\centering 
  \caption{Comparisons between our method with other published methods on the hold-out IRRG test images of Vaihingen Dataset.} 
\resizebox{\columnwidth}{!}{
\begin{threeparttable}
\begin{tabular}{c|p{9mm}|p{9mm}p{8mm}p{11mm}p{9mm}p{9mm}|p{8mm}} \hline
    \textbf{Models} & $\textbf{OA}$ & \textbf{Surface} & \textbf{Building} & \textbf{Low-veg} & \textbf{Tree} & \textbf{Car}  & \textbf{mF1} \\  \hline \hline
    UOA \cite{lin2016efficient}                 & 0.876  & 0.898  & 0.921  & 0.804 & 0.882  & 0.820  & 0.865 \\  
    DNN\_HCRF \cite{liu2019semantic}            & 0.878  & 0.901  & 0.932  & 0.814 & 0.872  & 0.720  & 0.848\\ 
    ADL\_3 \cite{paisitkriangkrai2015effective} & 0.880  & 0.895  & 0.932  & 0.823 & 0.882  & 0.633  & 0.833 \\ 
    DST\_2 \cite{Sherrah16}                     & 0.891  & 0.905  & 0.937  & 0.834 & 0.892  & 0.726 & 0.859 \\ 
    ONE\_7 \cite{audebert2016semantic}          & 0.898  & 0.910  & 0.945  & \textbf{0.844} & 0.899  & 0.778 & 0.875\\ 
    DLR\_9 \cite{MarmanisSWGDS16}               & 0.903  & 0.924  & 0.952  & 0.839 & 0.899  & 0.812  & 0.885 \\  
    GSN \cite{wang2017gated}                    & 0.903  & 0.922  & 0.951  & 0.837 & \textbf{0.899} & 0.824  & 0.887 \\ 
    DDCM-R50 \cite{liu_2019} & \textbf{0.904} & \textbf{0.927}& \textbf{0.953}    & 0.833   &0.894 & \textbf{0.883} & \textbf{0.898}  \\ \hdashline%
     $\textbf{SCG-GCN}$ & \textbf{0.904}  & 0.924  & 0.948  & \textbf{0.839} & \textbf{0.897}  & 0.880  & \textbf{0.898} \\  \hline
\end{tabular}
  \end{threeparttable}
} 
\label{tab:vaihingen_scores}%

\end{table}
\subsubsection{Comparison with other work}
We compare our results to other related published work on the ISPRS Potsdam RGB dataset and Vaihingen IRRG dataset. These results are shown in Table~\ref{tab:potsdam_scores} and~\ref{tab:vaihingen_scores} respectively. Our single model achieves an overall F1-score ($92.0\%$) on the Potsdam RGB dataset, which is comparable to state-of-the-art ($0.3\%$ point lower compared to the best model DDCM-R50~\cite{liu_2019}, but $0.2\% \sim 6.0\%$ higher than the remaining models). Similarly, our model trained on the Vaihingen IRRG images, also obtained very competitive performance with 89.8\% F1-score, which is around $+1.1\%$ higher than GSN~\cite{wang2017gated} and the same as the best model DDCM-R50. Fig.~\ref{fig:test} shows the qualitative comparisons of the land cover segmentation results from our model and the ground truths on the test set. 
\begin{table}[hptb!]
\centering 
  \caption{Quantitative Comparison of parameters size, FLOPs (measured on input image size of $3 \times 256 \times 256$), Inference time on CPU and GPU separately, and mIoU  on Potsdam RGB dataset.}
\resizebox{0.95\columnwidth}{!}{
\begin{threeparttable}
\begin{tabular}{c|c|p{13mm}p{11mm}p{21mm}p{11mm}} \hline 
\textbf{Models} & \textbf{Backbones} & \textbf{Parameters} \newline (Million)& \textbf{FLOPs} \newline (Giga) & \textbf{Inference time} \newline (ms - CPU/GPU) &\textbf{mIoU}$^{*}$  \\ \hline  \hline
U-Net \cite{ronneberger2015u}& VGG16 & 31.04 & 15.25 & 1460 / 6.37  & 0.715 \\
FCN8s \cite{long2015fully} & VGG16 & 134.30 & 73.46 & 6353 / 20.68 & 0.728 \\
SegNet \cite{badrinarayanan2017segnet}& VGG19 & 39.79 & 60.88 & 5757 / 15.47 & 0.781 \\
GCN \cite{peng2017large} & ResNet50 & 23.84 & 5.61  & 593 / 11.93 & 0.774 \\
PSPNet \cite{zhao2016pyramid} & ResNet50 & 46.59 & 44.40 & 2881 / 81.08 & 0.789 \\
DUC \cite{wang2017understanding} & ResNet50 & 30.59 & 32.26 & 2086 / 68.24 & 0.793 \\ DDCM-R50 \cite{liu_2019} & $\text{ResNet50}^\dagger$ & 9.99  & 4.86 & 238 / 10.23 &\text{0.808} \\
 \hdashline
$\textbf{SCG-GCN}$& $\text{ResNet50}^\dagger$ & \cellcolor{gray!25}\textbf{8.74}  &\cellcolor{gray!25}\textbf{4.47} & \cellcolor{gray!25}\textbf{161} / \textbf{6.08} & \cellcolor{gray!25}\textbf{0.810}  \\ 
\hline
\end{tabular}
\begin{tablenotes}
    \item[*] mIoU was measured on full reference ground truths of our local test images 5\_11, 6\_9 and 7\_11 in order to fairly compare with our previous work \cite{liuqinghui2018}.
\end{tablenotes}
\end{threeparttable}
} 
\label{tab:parameters}%

\end{table}

\subsubsection{Comparison of computational efficiency}
We also compared our methods to some popular architectures on the local Potsdam RGB test set~\cite{liuqinghui2018} in terms of parameter size, computational cost (FLOPs), inference time on both CPU and GPU, and mIoU evaluated on the full reference ground truths of the dataset. Table~\ref{tab:parameters} details the quantitative results of our SCG-Net against others. Compared to PSPNet~\cite{zhao2016pyramid} and SegNet~\cite{badrinarayanan2017segnet}, our model consumes about $10$x and $13$x less FLOPs with $5$x and $4.6$x fewer parameters and $21$x and $42$x faster inference speed on CPU, but achieves $2.1\%$ and $2.9\%$ higher mIoU respectively. In addition, compared with the best model DDCM-R50, our model has about $12\%$ fewer parameters with around $40\%$ faster inference speed on GPU, and achieves better performance on our local Potsdam test set.

\subsection{Analysis and ablations}
In our analysis and ablation studies, we explore how the graph size and other components of our framework such as the adaptive residual term and regularization terms affect the training and final performance. We also analyze the learned node graphs and point out some limitations of our model. For the study we choose the $\operatorname{SCG-GCN}$ model with ResNet50 backbone and train and evaluate it on the Vaihingen IRRG dataset. 
\begin{table*}[hptb!]
\centering 
  \caption{Test performance of different settings on Vaihingen test set.}
\resizebox{0.75\textwidth}{!}{
\begin{threeparttable}
\begin{tabular}{ccc|cc|cc|cc|cc|cc} \hline
    $\gamma\cdot \mathbf{\hat{Z}}$ & $\mathcal{L}_{kl}+\mathcal{L}_{dl}$ & $\operatorname{SCG}$ & \textbf{OA} &  $\Delta\%$& \textbf{Building} & $\Delta\%$ & \textbf{Car} & $\Delta\%$ & \textbf{mF1} & $\Delta\%$ & Steps (K)& $\Delta$\\  \hline  
    \xmark  & \xmark & \cmark  & 0.899 & -0.5  &0.946  & -0.2 &-0.877  & -0.3  & 0.893 & -0.5 & 97 & 4.1x\\ 
    \xmark  & \cmark & \cmark  & 0.899 & -0.5   &0.946& -0.2  & 0.860& -2.0  & 0.890 & -0.8 & 46 & 1.9x\\ 
    \cmark  & \xmark & \cmark  & 0.895 & -0.9  &0.941  & -0.7 &0.878  & -0.2  & 0.891 & -0.7 & 15 & 0.6x\\ 
    \cmark  & \cmark & \cellcolor{gray!25}\xmark  & 0.887 & \cellcolor{gray!25}-1.7  &0.929  & \cellcolor{gray!25}-1.9 &0.829  & \cellcolor{gray!25}-5.1  & 0.874 & \cellcolor{gray!25}-2.4 & 120 & \cellcolor{gray!25}5.0x\\ 
    \cmark  & \cmark & \cmark  & \textbf{0.904} & -  & \textbf{0.948} &-  & \textbf{0.880} &-  & \textbf{0.898} & - & \textbf{24} & -\\\hline
\end{tabular}
\begin{tablenotes}
            \item[*] Steps (K) denote training iterations with K=$1000$. Note that 1K steps = 1 epoch in this work.
        \end{tablenotes}
  \end{threeparttable}
} 
\label{tab:ablation_adaptive}%

\end{table*}

\subsubsection{Effect of the adaptive residual and regularization terms}
\begin{figure}[htpb!]
 \centering
  \includegraphics[width=\columnwidth]{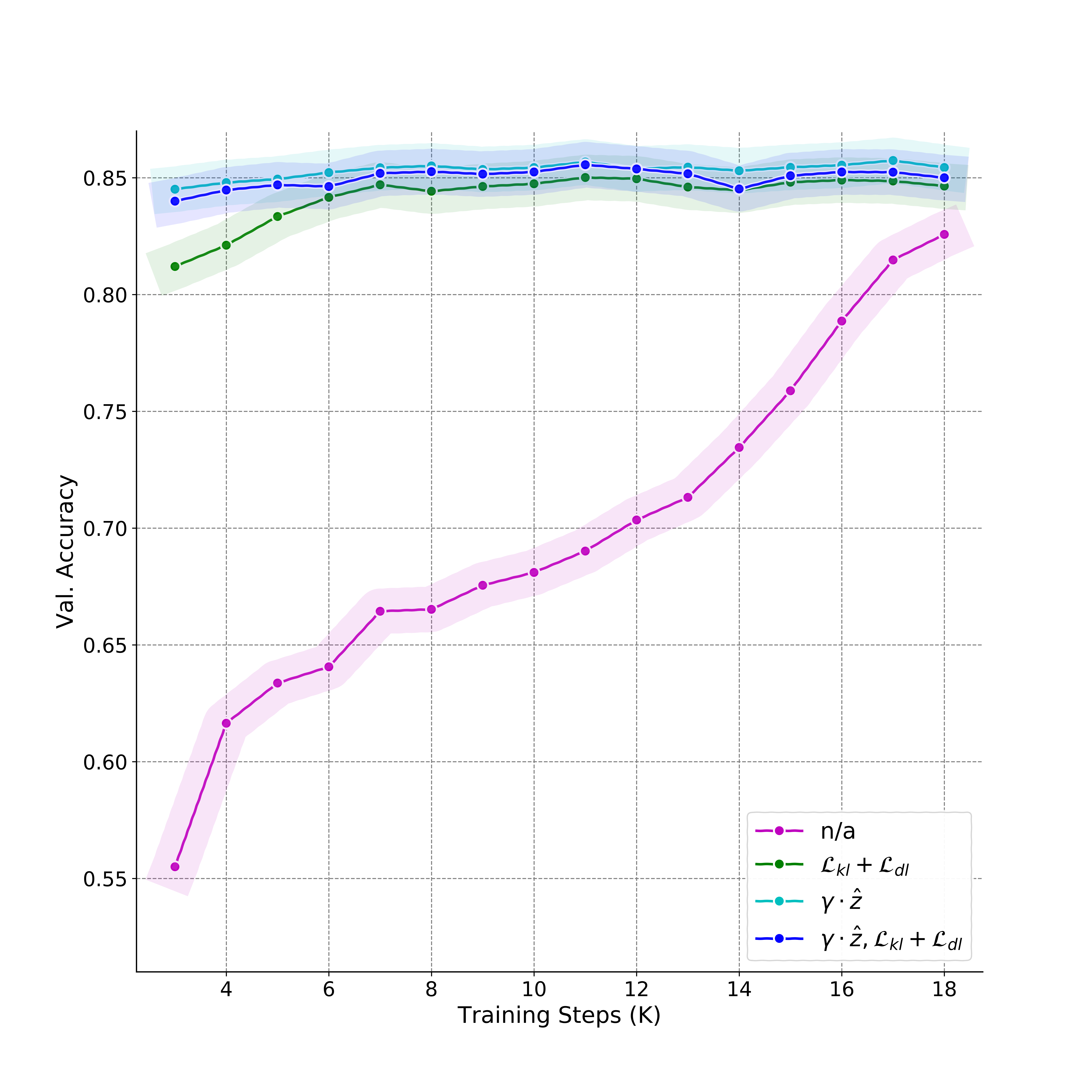} 
  \caption{Training performance of different settings. Here K means $1000$ and n/a denotes the training without using adaptive residual terms ($\gamma\cdot \mathbf{\hat{Z}}$) and regularization ($\mathcal{L}_{kl}+\mathcal{L}_{dl}$).}
  \label{fig:ablation_fig}
\end{figure}

We evaluated the effectiveness of the proposed adaptive residual term and the regularization. The test performance is presented in Table~\ref{tab:ablation_adaptive}. Generally, without using adaptive residual terms ($\gamma\cdot\mathbf{\hat{Z}}$) and regularization ($\mathcal{L}_{kl}+\mathcal{L}_{dl}$), the number of training steps required to reach convergence is increased by $4.1$x, and when regularization is used alone, the training converges faster, but the test performance is significantly decreased on small objects (i.e, cars $-2.0\%$) and overall ($-0.8\%$). Similarly, when only applying the residual term, the convergence speed was even faster than the default setting ($0.6$x training steps) while the test results also became worse both on big objects like buildings ($-0.5\%$) and overall classes ($-0.7\%$). Fig.~\ref{fig:ablation_fig} illustrates the training performance of the first $18$ training epochs with different settings. Note, both the residual term and the regularization loss can stabilize the training process and speed up the convergence. However, when the residual term or the regularization loss is used alone, we observe some degradation in performance. 
\begin{table*}[hptb!]
\centering 
  \caption{Test performance of different input sizes and node sizes on Vaihingen test set.}
\resizebox{0.8\textwidth}{!}{
\begin{threeparttable}
\begin{tabular}{cccc|cc|cc|cc|cc} \hline
    Input  & Nodes & GFLOPs & FPS & \textbf{OA} &  $\Delta\%$& \textbf{Building} & $\Delta\%$ & \textbf{Car} & $\Delta\%$ & \textbf{mF1} & $\Delta\%$\\  \hline     
    $256\times 256\times 3$ & $16^2$  & 4.37  & 178   & 0.903 & -0.1  &0.944  & -0.4 &-0.858  & -2.2  & 0.893 & -0.5\\   
     $512\times 512\times 3$ & $32^2$  & 17.5  & 91   & 0.901 & -0.3  &0.948  & - &0.875  & -0.5  & 0.894 & -0.4\\
     $768\times 768\times 3$ & $48^2$ & 39.4  & 44   & 0.901 & -0.3  &0.946  & -0.2 &0.870  & -1.0  & 0.893 & -0.5\\  \hline  
     $448\times 448\times 3$ & $28^2$  & 13.4  & 113   & \textbf{0.904} & -  & \textbf{0.948} &-  & \textbf{0.880} &-  & \textbf{0.898} & -\\\hline
\end{tabular}
\begin{tablenotes}
            \item[*] FPS means frames per second on GPU (i.e., NVIDIA 1080Ti GPU in this work).
        \end{tablenotes}
  \end{threeparttable}
} 
\label{tab:ablation_nodes}%

\end{table*}

\subsubsection{Effect of the graph size}
We investigated the effect of node size and corresponding input image size on the performance. Table~\ref{tab:ablation_nodes} presents the details of the evaluation results where four models are trained on various input and node settings. Smaller graph size ($\text{nodes} = 16^2$) and input size ($256\times 256\times 3$) can lead to very fast inference speed but also result in significantly worse accuracy on small objects such as cars. Larger node size ($\text{nodes} = 32^2 \text{or } 48^2$ considerably slows down the inference speed without obtaining better performance ($-0.4\%$ in terms of mF1-score). So overall, our model with node size of $28^2$ achieves the best results on the Vaihingen dataset with very fast inference speed ($113$ FPS on a GPU). Note that the node size has no effect on the number of parameters of the model. 

\subsubsection{The learned graphs}
\begin{figure*}[htpb!]
 \centering
  \includegraphics[width=\textwidth]{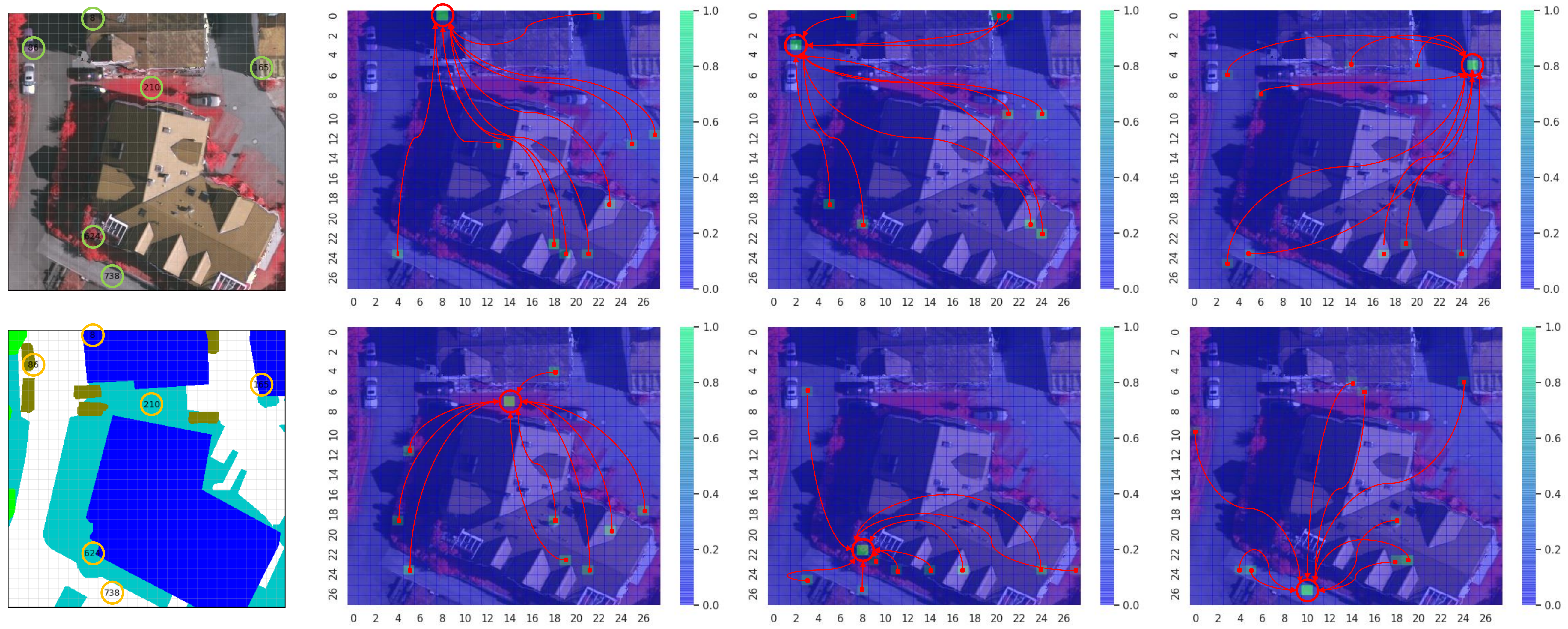} 
  \caption{Visualization of the learned node graphs ($28^2 \text{ nodes}$) onto 2D relation maps. From left to the right and top to bottom, the relation maps of node-8, 86, 165, 210, 624 and 738 are shown respectively. Dark blue regions represent weak dependency to the corresponding node, while light color blocks indicate strong relations to the node. Note that we normalized the edge weights to $[0.0, 1.0]$ and only illustrated the top nine largest weights associated to the target node.}
  \label{fig:node_graph}
\end{figure*}

We consider that the effectiveness and efficiency of our SCG-Net is mainly relying on the proposed SCG module that is able to construct the underlying non-local contextual relations in an end-to-end learnable manner. We visualized the SCG modules learned node graphs as 2D relation maps (see Figure~\ref{fig:node_graph}). We highlight 6 representative nodes/squares (labeled with 8, 86, 165, 210, 624 and 738 separately) marked with 6 circles on both the input image and the ground truth. There are 6 relation maps from left to the right and top to bottom representing the learned relation graphs for the six nodes respectively, where dark blue regions represent weak dependency to the corresponding node, while light color blocks indicate strong relations to the target node. The relation-map visualizes the learned top-9 weighted-relationships of the target node (circled square patch) w.r.t its global contextual nodes (light-colored patches in the images), with gradient color palettes from light green to dark blue indicating the transition from strong dependency to zero-dependency. We observe that the target node has been able to interact with long-range neighbor nodes via the learned relation graph. In addition, when the SCG (the learned node graph) is discarded from the model as shown in Table~\ref{tab:ablation_adaptive}, the performance of the model is greatly reduced (e.g. on Car $-5.1\%$ and mF1 $-2.4\%$), and also the number of training steps dramatically increases by 5 times. We thus consider that the learned graphs can benefit semantic segmentation tasks in terms of training process and performance, but believe that future studies are required to enhance the interpretability of the learned dependencies.
\begin{table}[hptb!]
\centering 
  \caption{Comparisons of SCG-Net variants on the hold-out IRRG test images of Vaihingen Dataset.} 
  
\resizebox{\columnwidth}{!}{
\begin{threeparttable}
\begin{tabular}{c|c|p{9mm}p{8mm}p{11mm}p{9mm}p{9mm}|c} \hline
    \textbf{SCG-Net Variants} & $\textbf{OA}$ & \textbf{Surface} & \textbf{Building} & \textbf{Low-veg} & \textbf{Tree} & \textbf{Car}  & \textbf{mF1} \\  \hline \hline
    $\operatorname{SCG-GIN}$ & 0.901  & 0.922  & 0.947  & 0.836 & 0.894  & 0.877  & 0.895 \\  
    $\operatorname{SCG-GCN-GIN}$ & 0.899  & 0.920  & 0.946  & 0.828 & 0.893  & \textbf{0.888}  & 0.895 \\  
    $\operatorname{SCG_{ae}-GCN}$ & 0.897  & 0.919  & 0.942  & 0.827 & 0.892  & 0.880  & 0.892 \\  
    $\operatorname{SCG_{dir}-GCN}$ & 0.902  & 0.923  & 0.947  & 0.833 & 0.895  & 0.881  & 0.896 \\  
    $\operatorname{SCG_{dir}-GCN-GIN}$ & 0.900  & 0.920  & 0.946  & 0.832 & 0.895  & 0.886  & 0.896 \\   
    $\operatorname{SCG-GCN_{\_ns}}$ & 0.899  & 0.919  & 0.946  & 0.834 & 0.892  & 0.860  & 0.890 \\  \hdashline
     $\textbf{SCG-GCN}$ & \textbf{0.904}  &\textbf{0.924}  & \textbf{0.948}  & \textbf{0.839} & \textbf{0.897}  & 0.880  & \textbf{0.898} \\  \hline
\end{tabular}
  \end{threeparttable}
} 
\label{tab:variants_scores}%

\end{table}

\subsubsection{SCG-Net variants}
Additionally, we evaluated variants of the SCG-Net with different configuration settings as shown in Table~\ref{tab:models}. All models except $\operatorname{{SCG}_{ae}-GCN}$ and $\operatorname{{SCG}-GCN_{\_ns}}$ demonstrated very close performance ($\Delta \sim \pm 0.3\%$ in terms of mF1-score) on the test set of Vaihingen dataset as shown in Table~\ref{tab:variants_scores}, and we further observe that when GCN and GIN are combined together in the framework, the models, such as $\operatorname{{SCG}-GCN-GIN}$ and $\operatorname{{SCG}_{dir}-GCN-GIN}$, obtained considerably better results on the small object (i.e, cars $+0.6\sim0.8\%$ ).
We therefore hypothesize that combining spectral-based graph neural networks (such as GCN) with spatial-based graph neural networks (such as GIN) can improve propagation of information and refine small object predictions. How to efficiently utilize this observation will be explored in future work.

\subsubsection{Limitations}
\begin{figure*}[htpb!]
 \centering
  \includegraphics[width=\textwidth]{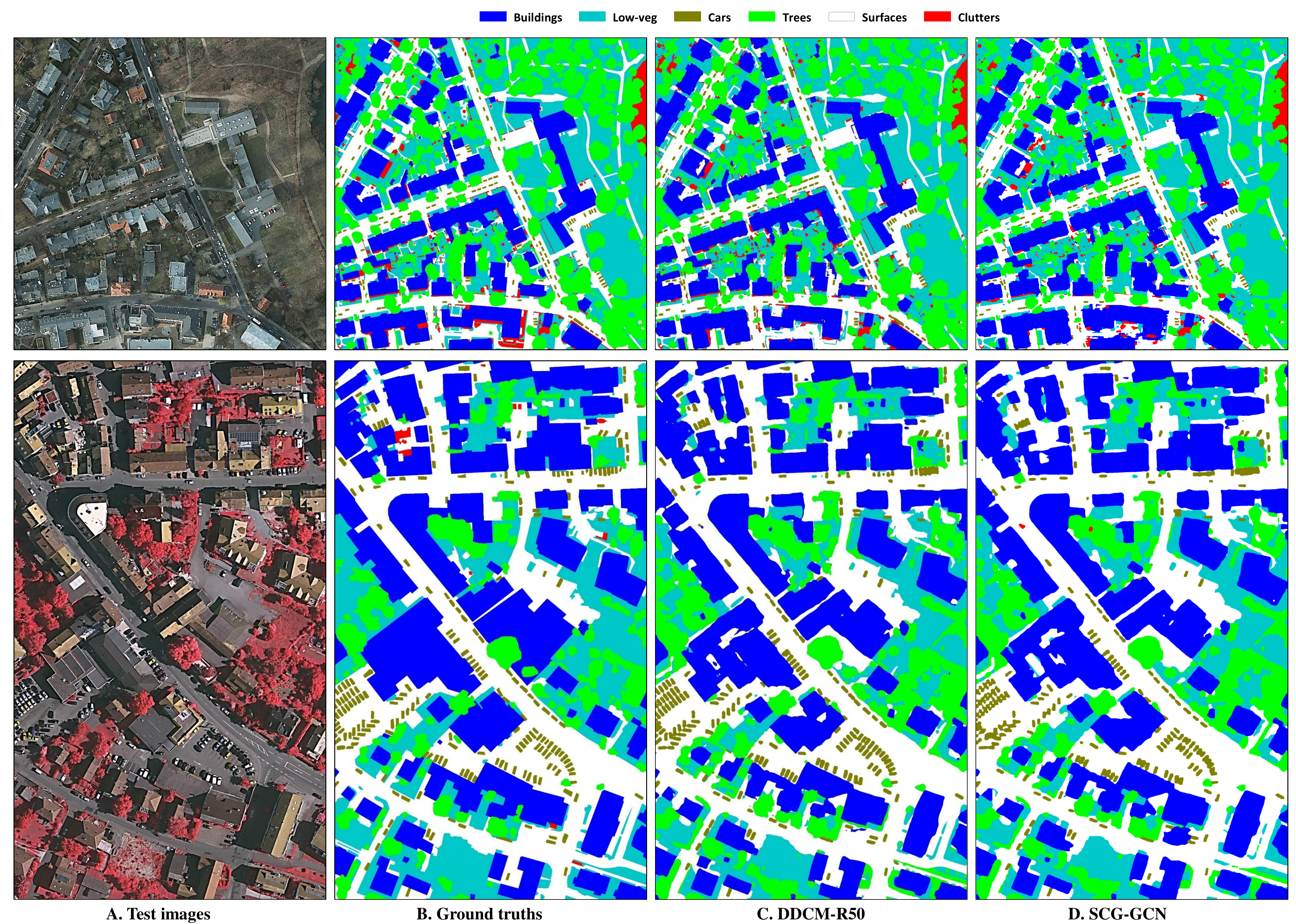}
  \caption{Segmentation results for test images of Potsdam tile-3\_14 (top) and Vaihingen tile-27 (bottom). From the left to right, the input images, the ground truths and the predictions of DDCM-R50, and our SCG-Net.}
  \label{fig:test}
\end{figure*}

The performance of GNNs is known to gradually decrease with increasing number of layers partly due to its over-smoothing issue, where repeatedly applying graph convolutions eventually makes features of vertices indistinguishable. Our model therefore exploits only 2-layer GNNs as the decoder and final semantic prediction simultaneously. Despite its promising performance, stacking more GNN layers either in the decoder or for the prediction significantly hurts the training and test performance of our model. In addition, we observed that segmentation performance on the boundaries of small and dense objects (e.g, cars) was not as good as the baseline DDCM model. The closely located small objects tend to be segmented as a whole big object (see Figure~\ref{fig:test}). We consider that the existing GNN architectures have limited capability in capturing and decoding the spatial location of a given node with respect to all other nodes. We plan future work to address these limitations.

\section{Conclusions} \label{concl}
We presented a novel self-constructing graph (SCG) framework, which makes use of learnable latent variables to effectively construct the global context relations (latent graph structure) directly from the input feature maps. SCG can obtain long-range dependencies from complex-shaped objects without relying on manually built prior knowledge graphs. Built upon SCG and graph neural networks (GNNs), we developed a new end-to-end method (SCG-Net) that streamlines semantic segmentation as a node classification problem. Our SCG-Net achieves competitive performance on the publicly available ISPRS Potsdam and Vaihingen datasets, with much fewer parameters, and at a lower computational cost compared to strong baseline models. 



\section*{Acknowledgment}
This work is supported by the foundation of the Research Council of Norway under Grant 220832. 


\ifCLASSOPTIONcaptionsoff
  \newpage
\fi



%

\bibliographystyle{IEEEtran}
\bibliography{bibtex/my.bib}

\end{document}